\documentclass[11pt]{article}

\usepackage[preprint]{acl}

\usepackage{times}
\usepackage{latexsym}

\usepackage[T1]{fontenc}

\usepackage[utf8]{inputenc}

\usepackage{microtype}

\usepackage{inconsolata}

\usepackage{graphicx}
\usepackage{amsmath}

\usepackage{todonotes}
\usepackage[shortlabels,inline]{enumitem}
\setlist[itemize]{nosep,leftmargin=*}
\usepackage{cleveref}
\crefname{section}{s}{Sections}
\crefname{table}{Table}{Tables}
\crefname{figure}{Fig.}{Figs.}
\crefname{algorithm}{Alg.}{}
\crefname{ALC@unique}{Line}{Lines}
\crefname{equation}{Eq.}{Eqns.}
\crefname{appendix}{Appendix}{}
\crefformat{section}{\S#2#1#3}
\usepackage[normalem]{ulem}
\usepackage{graphicx}

\definecolor{darkred}{RGB}{180,0,0}

\definecolor{darkgreen}{RGB}{0,150,0}

\usepackage{graphicx}
\usepackage[most]{tcolorbox}
\usepackage{multirow}
\usepackage{tcolorbox}

\PassOptionsToPackage{table}{xcolor}
\usepackage{xcolor}      

\usepackage{wrapfig}
\usepackage{floatflt}
\usepackage{color, colortbl}

\usepackage[inline,shortlabels]{enumitem}

\usepackage{lineno}

\definecolor{darkblue}{rgb}{0, 0, 0.5}
\hypersetup{colorlinks=true, citecolor=darkblue, linkcolor=darkblue, urlcolor=darkblue}
\definecolor{Highlight}{rgb}{0.92,0.94,1}

\usepackage{tabularx}
\PassOptionsToPackage{table}{xcolor}
\usepackage{booktabs}
\usepackage{wrapfig}
\usepackage{graphicx}
\usepackage{amsfonts} 
\usepackage{float}
\usepackage{color, colortbl}

\title{SCRIBE: Structured Mid-Level Supervision for Tool-Using Language Models}

\author{
Yuxuan Jiang$^{1}$ \quad Francis Ferraro$^{1}$ \\
$^{1}$University of Maryland, Baltimore County \\
\texttt{yuxuanj1@umbc.edu}
}

\begin{document}
\maketitle

\begin{abstract}
Training reliable tool-augmented agents remains a significant challenge due to the difficulty of \textbf{credit assignment} in multi-step reasoning. While Process-level Reward Models (PRMs) offer a potential solution, standard LLM-based judges often provide inconsistent signals because they lack granular, task-specific rubrics to disentangle high-level planning from low-level execution. In this work, we propose \textbf{SCRIBE} (\textbf{S}kill-\textbf{C}onditioned \textbf{R}eward with \textbf{I}ntermediate \textbf{B}ehavioral \textbf{E}valuation), a reinforcement learning framework that intervenes at a novel \textbf{mid-level abstraction}. SCRIBE anchors reward modeling in a curated library of \textit{Skill Prototypes}, transforming open-ended LLM evaluation into a constrained verification task. By routing subgoals to specific prototypes, we provide the judge with precise rubrics that significantly reduce reward variance. Empirical results demonstrate that SCRIBE achieves state-of-the-art performance across reasoning and tool-use benchmarks; notably, it improves the \textbf{AIME25} score of a Qwen3-4B model from \textbf{43.3\% to 63.3\%} and substantially enhances success rates in complex multi-turn tool interactions. Furthermore, our analysis of training dynamics characterizes a \textbf{co-evolution} between levels, where mid-level skill mastery serves as a precursor to the emergence of strategic high-level planning. Finally, we show that SCRIBE is \textbf{additive} to low-level tool optimizations, offering a scalable and complementary approach to building more autonomous and reliable agents.
\end{abstract}

\section{Introduction}
Tool-integrated reasoning augments the reasoning process with external tool invocations, providing verifiable signals and the potential to improve a model’s performance ceiling comparing to traditional text-only reasoning~\cite{yao2023react,qian2025toolrl}. However, the large and diverse space of available tools makes learning effective tool use challenging~\cite{xue2025simpletir,liu2024toolace}. Models often fail to achieve intermediate reasoning objectives due to unreliable tool selection, invocation, or result integration, limiting the benefits of tool augmentation. As a result, enabling models to reliably use tools remains a central challenge in tool-augmented reasoning.
\begin{figure*}
    \centering
    \includegraphics[width=1\linewidth]{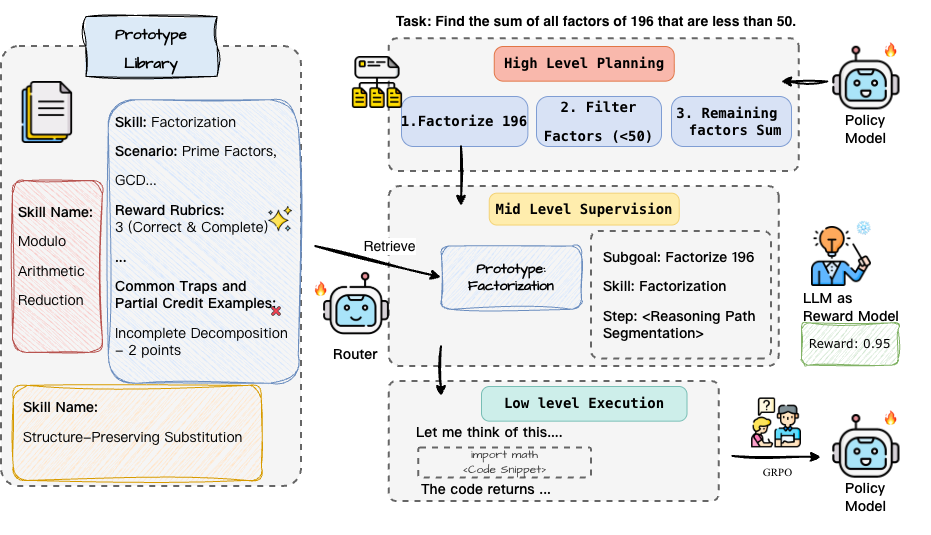}
    \caption{Overview of our three-stage framework. The policy model performs high-level planning, mid-level reasoning, and low-level execution. At the mid level, a router retrieves skill-specific prototypes from a prototype library and uses them to structure reward evaluation by a frozen LLM-based (GPT5-mini) reward model. The policy is optimized with GRPO using the resulting rewards. Flame denotes trainable components, while snowflake indicates frozen models.}

    \label{fig:intro}
\end{figure*}
Recent advances in agentic reinforcement learning have demonstrated promising gains in tool-augmented reasoning tasks~\cite{zhang2025landscape}.
However, training tool-using agents remains challenging due to credit assignment.
In multi-step and multi-tool settings, outcome-level rewards are often insufficient, as errors may arise from planning, execution, or improper tool use, which are difficult to disentangle from final outcomes alone~\cite{qian2025toolrl}.
To mitigate this issue, prior work has explored process-level reward modeling (PRM), typically relying on LLM-based judges to score intermediate reasoning steps or full trajectories~\cite{yu2025demystifying,zhang2025lessons,khalifa2025process}.
Despite scalability, such judges often yield inconsistent signals due to underspecified reward criteria.

For example, in a multi-step mathematical problem, an LLM-based judge may assign a high score to a reasoning trajectory simply because the final numerical answer is correct, while overlooking logical flaws in intermediate tool usage, such as misinterpreting an API response or relying on an unjustified computational shortcut.
Conversely, a step may be penalized due to a minor, non-fatal syntax error in a tool call, even when the underlying reasoning strategy is sound and the error does not affect the final outcome.
Without a rigorous rubric that distinguishes between strategic planning and technical execution, such inconsistent reward signals obscure which intermediate decisions actually contributed to success or failure, hindering effective learning during training.

While prior work often oscillates between supervising high-level plans or low-level tool execution, we argue that the \textbf{mid-level abstraction} provides a more effective point of intervention for robust credit assignment. As illustrated in Figure~\ref{fig:intro}, we propose \textbf{SCRIBE} (Skill-Conditioned Reward with Intermediate Behavioral Evaluation), a framework that provides structured supervision for tool-augmented reasoning .
SCRIBE centers on a curated library of \emph{Skill Prototypes}, each representing a canonical mid-level reasoning pattern distilled from clusters of semantically similar subgoals and skills.
An auxiliary \emph{Router} maps raw reasoning trajectories into structured $\langle \text{subgoal, skill, step} \rangle$ representations and associates each subgoal with its corresponding prototype. 
Crucially, this taxonomic anchoring enables a more reliable reward scheme by grounding open-ended LLM judgments in structured constraints. Rather than tasking a judge with evaluating arbitrary reasoning steps—a process often prone to "hallucinatory" rewards or inconsistent criteria—SCRIBE provides the judge with a context-specific rubric defined by the routed Skill Prototype. By narrowing the evaluation space from subjective logical assessment to the verification of specific skill execution, we effectively transform the reward modeling from an open-ended generation task into a constrained verification task. This significantly reduces reward variance and ensures that the reinforcement learning signal is both dense and grounded in specific reasoning behaviors.

We evaluate \textsc{SCRIBE} on challenging benchmarks for mathematical reasoning (MATH, AIME25) and tool-use (BFCL V4). As shown in Table~\ref{tab:main_results}, \textsc{SCRIBE} consistently outperforms state-of-the-art RL baselines. Notably, on the \textbf{AIME25} benchmark, our approach improves the Qwen3-4B-instruct-2507 model from 43.3\% to \textbf{63.3\%}. In complex tool-use scenarios (\textbf{BFCL Multi}), \textsc{SCRIBE} achieves a \textbf{33.3\%} success rate, demonstrating its superior ability to handle multi-step tool interactions through precise credit assignment.

Beyond performance gains, our analysis of training dynamics (Sec.~\ref{sec:rq1_dynamics}) reveals a co-evolution between mid-level execution and high-level planning. As shown in Figure~\ref{fig:ab1_structural_dynamics}, the mastery of intermediate skills acts as a precursor to the emergence of strategic coherence. These results indicate that mid-level supervision does not merely improve local execution fidelity, but fundamentally reshapes the model’s high-level ability by providing a more stable foundation for long-horizon planning.

In summary, our contributions are as follows:
\begin{itemize}
    \item We formalize a three-level hierarchy for tool-use, identifying mid-level skills as the optimal point for credit assignment.
    \item We introduce \textsc{SCRIBE}, which anchors rewards in \textit{Skill Prototypes} to transform noisy evaluation into grounded verification.
    \item We get empirical performance gain on results and characterize a \textit{co-evolution} where mid-level mastery facilitates high-level planning.
    \item We demonstrate that \textsc{SCRIBE} complements low-level tool optimization, yielding superior synergistic performance in Sec.~\ref{ab2}.
\end{itemize}

\section{Related Works}
\subsection{Tool-Integrated Reasoning with Agentic Reinforcement Learning}
Among various LLM abilities, foundational research in semantic matching~\cite{xue2023dual, xue2025structcoh} and temporal QA~\cite{xue2024qa} has paved the way for complex understanding. While domain adaptation~\cite{huang2023dtbs} and test-time integration~\cite{huang2025cosmic} enhance robustness, reasoning ability achieved by post-training~\cite{huang2026rlvr, lan2025mappo} stands as a key frontier. Given that traditional SFT may suffer from incomplete learning~\cite{xue2026sft}, recent works introduce uncertainty-aware reward modeling~\cite{xue2026reward}, dense feedback from LM critics~\cite{cao2024enhancing}, and preference-driven code generation~\cite{li2025preference} to refine alignment. Moreover, recent studies have advanced role-playing reasoning through human-actor emulation~\cite{chen2026actormindemulatinghumanactor} and improved continual learning via prototype-conditioned generative replay~\cite{chen-zeng-2025-prototype}.

The scope of reasoning has expanded into multimodal domains, where models now "think with 3D"~\cite{chen2025think} and utilize reinforced visual perception~\cite{chen2025visrl, chen2025sifthinker}. Advanced frameworks like OmniVideo-R1~\cite{chen2026omnivideo} and DV-Matcher~\cite{chen2025dv} further integrate audio-visual and geometric cues. Such multimodal intelligence significantly boosts retrieval performance, including chat-driven search~\cite{xie2025chat, xie2026conquer} and vision-driven video representation~\cite{xie2026hvd, xie2026delving}. Furthermore, safeguarding techniques~\cite{yu2024robust, lan2025contextual} and hallucination mitigation~\cite{yu2024mechanistic, ji2025calibrating} ensure the reliability of these complex systems.

To support deployment, system efficiency and continual learning have become indispensable. Inference is optimized via multi-tier KV cache management~\cite{chu2025mcam}, selective sharing~\cite{chu2025selective}, and adaptive MoE scheduling~\cite{chu2025dynamic, shen2025expertflow}. Moreover, lightweight techniques like parameter-efficient sampling~\cite{yao2024swift}, structured pruning~\cite{li2026sepprune}, efficienct context management~\cite{shen2026efficiency} and distillation~\cite{li2025frequency} enable efficient task execution, from medical imaging~\cite{Li2025Efficient} to web agents~\cite{ding2026dynaweb}. Finally, reversible alignment~\cite{xiao2026reversible} and unified LoRA generation~\cite{xiaometa, chen2024three} allow models to learn continually, while comprehensive interaction surveys~\cite{li2026comprehensive} and robust federated learning~\cite{chen2024confusion} continue to push the boundaries of scalable AI.

Beyond core architecture, recent research emphasizes a data-centric perspective on the entire LLM lifecycle~\cite{rao2025data}, where mathematical reasoning is further enhanced by adversarial data synthesis~\cite{yu2026mathagent} and iterative DPO with dynamic sampling~\cite{rao2025dynamic}. To improve model robustness and learning efficiency, advanced techniques such as difficulty-based sample weighting~\cite{zhou2022understanding}, implicit data augmentation~\cite{zhou2024boosting}, and adversarial training with anti-adversaries~\cite{zhou2024adversarial} have been introduced.

In the context of embodied AI and spatio-temporal reasoning, unified token spaces for human-object interaction~\cite{yang2026unihoi} and object-centric decoupling for robotic manipulation~\cite{yang2026instrucrobo} have improved explainable decision-making. These abilities are further integrated into 3D scene understanding~\cite{yang2026unibvr} and autonomous driving through spatio-temporal Chain-of-Thought (CoT)~\cite{zeng2025fsdrive} and dual-memory vision-language navigation~\cite{zeng2025janusvln}. Furthermore, the application of multi-scale graph learning and Transformers has shown significant potential in predicting complex real-world dynamics, including financial fraud detection~\cite{lei2026multiscale}, traffic flow prediction~\cite{shen2026mftformer}, disaster risk assessment~\cite{shen2025aienhanced}, mental health~\cite{yang2026exploring}, recommendations~\cite{wang2025improving} and socio-economic metrics such as labor market gradients~\cite{liu2026health} and business resilience indices~\cite{Shen2026bri}.

Tool-integrated reasoning extends large language models by enabling them to invoke external tools such as code interpreters and search engines, substantially improving the limits of purely textual reasoning~\cite{yao2023react, wang2024mathcoder, liao2024mario}. This paradigm builds upon the diverse foundational capabilities of LLMs in reasoning, knowledge retrieval, and instruction following~\cite{jiang2024memorization,zhang2025find,li2025faedkvinfinitewindowfouriertransform,jiang2026cornerstonesstumblingblocksdeciphering}.
Early tool-use frameworks such as ToRA~\cite{gou2023tora} and Toolformer~\cite{schick2023toolformer} rely primarily on supervised fine-tuning to teach models when and how to invoke external tools.
Building on this line, a series of recent \emph{agentic reinforcement learning} approaches ,likeToolRL~\cite{qian2025toolrl}, DeepSeek-R1~\cite{guo2025deepseek-r1}, explicitly model tool usage as actions and optimize multi-step tool interaction via reward design, such as DAPO~\cite{yu2025dapo} and GRPO~\cite{liu2025Dr.GRPO}.

Despite their success, most existing methods concentrate on \emph{low-level} tool-use behaviors, focusing on richer invocation patterns, improved execution accuracy, or increased autonomy at the action level~\cite{dong2025toolstar, xu2025learning, li2025one}, rather than evaluating or supervising plan-level discrimination and high-level decision structure. In contrast, it remains underexplored whether tool use improves \emph{mid-level} reasoning quality, such as subgoal execution and intermediate abstraction, or translates into stronger \emph{high-level} reasoning.
Our work addresses this gap by examining how agentic tool use affects reasoning competence beyond surface-level tool invocation.

\subsection{Progress Reward Modeling with LLM-as-Judge}

Progress reward modeling extends outcome-based reward modeling by evaluating the correctness of intermediate reasoning steps rather than final answer only, thereby providing denser supervision for improving reasoning performance~\cite{lightman2023let, setlur2024rewarding}.
Recent work increasingly adopts LLM-as-judge frameworks to support PRM, leveraging their low annotation cost and strong generalization to produce large-scale supervision signals for agentic reinforcement learning~\cite{tan2024large, li2025generation, choudhury2025process,}.

Despite these advantages, LLM-based judges are known to suffer from vulnerability and inconsistency, which can induce reward hacking and destabilize training~\cite{zhao2025one, agarwal2025toolrm,li2025preference,li2025s}.
Such issues arise when models exploit weaknesses in the reward signal rather than improving genuine reasoning quality.
Our work explicitly analyzes this failure mode and proposes a more consistent and accurate evaluation protocol, leading to improved stability and performance.

\subsection{Problem Decomposition and skills in Reasoning}

Decomposing complex reasoning problems into a sequence of simpler subgoals has been shown to improve performance on challenging tasks~\cite{xiang2025can,teng2025atom,shi2022skill}.
With an appropriate decomposition, each subgoal typically corresponds to a dominant reasoning skill, where a skill denotes a reusable reasoning operation, such as algebraic factorization or querying external information~\cite{dalal2024plan}.

Prior work has shown that skills form a stable and model-recognizable level of structure within LLMs~\cite{jiang2025drp}.
Compared to step-level decomposition, skill-based decomposition enables more consistent identification of reusable reasoning patterns across problems.
Building on this insight, our work investigates how strengthening tool-related skills can improve subgoal execution and, consequently, overall reasoning performance.

\section{Method: Decomposition, Clustering, prototype}

Our framework, \textsc{SCRIBE}, decomposes the training process into three core stages: skill-level abstraction, structured reward evaluation, and policy optimization. As illustrated in the system overview (\textbf{Figure~\ref{fig:intro}}), we move beyond raw process-level rewards by introducing a mid-level "semantic bridge." We first distill reasoning trajectories into a library of reusable \emph{Skill Prototypes}. During training, an auxiliary \emph{Router} maps the student's rollouts to these prototypes, allowing an LLM-based judge to provide rewards that are calibrated against a consistent taxonomic rubric rather than subjective inference. Finally, the model is optimized using GRPO, benefiting from the densified and stabilized credit assignment signals.

\subsection{Mid-Level Skill Abstraction}
\label{prototype and cluster}
We formalize a trajectory as an ordered sequence of $\langle \text{subgoal, skill, step} \rangle$ triples, where each step represents a non-overlapping span of the reasoning trace. We show a simplified verision of prompt here and the full prompt is provided in Appendix~\ref{full prompt}. Our \textsc{SCRIBE} targets this mid-level abstraction to bridge the gap between high-level planning and low-level execution.

\tcbset{
  promptbox/.style={
    colback=gray!5,
    colframe=gray!50,
    boxrule=0.6pt,
    arc=2pt,
    left=6pt,right=6pt,top=6pt,bottom=6pt,
    fonttitle=\bfseries,
    breakable
  }
}
\begin{tcolorbox}[promptbox, title=Simplified Subgoal and Skill Extraction Prompt]
Given a problem or a tool-using trajectory, decompose the solution into a sequence of mid-level subgoals, where completing these subgoals in order is sufficient to solve the whole problem.
For each subgoal, identify:
(1) what the subgoal accomplishes,
(2) the primary reasoning skill required,
(3) a representative step or span where the skill is applied.
Example:
\noindent
\textbf{subgoal:} maximize $u = (x+4)y$ subject to $x^2 + y^2 = 1$ \\
\textbf{skill:} optimization with constraint (Lagrange multipliers or trigonometric substitution) \\
\textbf{step:} set $x = \cos \theta$, $y = \sin \theta$, so
\[
u(\theta) = (\cos \theta + 4)\sin \theta
           = \tfrac{1}{2}\sin 2\theta + 4\sin \theta,
\]
and maximize over $\theta$.
\label{prompt_ex}
\end{tcolorbox}

\paragraph{Prototype Construction.}To build this abstraction, we first decompose trajectories into subgoals and associated skills using the prompting scheme in Appendix A. We then apply a hierarchical clustering approach—utilizing HDBSCAN for dense clusters and K-means as a fallback—to group semantically similar reasoning patterns. For each cluster, we distill a representative \textit{Skill Prototype} that aggregates usage contexts, intermediate objectives, and canonical reasoning patterns while abstracting away instance-specific details like concrete numerical values.\paragraph{Structured Verification via Prototypes.}Skill Prototypes transform subjective, open-ended judging into structured verification. As shown in the Bound-Based Conclusion example (Figure~\ref{fig:skill_prototype_compact}), once a step is routed to a prototype, the LLM judge is provided with a precise, checklist-style rubric. For instance, instead of vaguely rewarding "logical correctness," the prototype requires verifying if the agent properly addressed inequality strictness or identified necessary bounds. Crucially, each prototype defines Common Traps—such as a "Boundary Leak" (e.g., failing to discretize a continuous bound like $c < 44.75$ to $c \le 44$)—and maps them to specific penalty scores. This granularity ensures the reward signal is a diagnostic reflection of skill mastery rather than a biased estimate based solely on the final outcome.

\begin{figure}[t]
\centering
\begin{tcolorbox}[
    enhanced,
    colback=gray!5,
    colframe=gray!70!black,
    fonttitle=\bfseries,
    title={Skill Prototype (Compact): Bound-Based Conclusion \& Synthesis},
    width=\linewidth,
    sharp corners
]
\setlength{\baselineskip}{0.95\baselineskip}
\setlength{\parskip}{0pt}

\textbf{Skill:} Bound-Based Conclusion and Synthesis. \;
\textbf{Use when:} a subgoal \emph{terminates} a reasoning chain by concluding from established bounds or constraints.

\vspace{2pt}
\textbf{Pattern:}
(i) collect relevant bounds $\rightarrow$
(ii) determine whether tightness/feasibility must be addressed $\rightarrow$
(see Appendix for remaining steps).

\vspace{2pt}
\textbf{Scoring Rubric (summary):}
\textbf{3} rigorous conclusion with tightness;
\textbf{2} correct conclusion with minor slip;
\textbf{1} major logical gap;
\textbf{0} wrong or premature conclusion.

\vspace{2pt}
\textbf{Common Trap (example):}
\emph{Boundary leak} — failing to discretize a continuous bound
(e.g., concluding $c < 44.75$ without stating $c \le 44$),
which leaves the subgoal incomplete (\textbf{Score = 2}).

\vspace{1pt}
\textbf{Note:}
Additional traps (implicit tightness, domain neglect, premature synthesis)
and the full scoring protocol are provided in the appendix.

\end{tcolorbox}
\caption{A compact illustration of a skill prototype used for cluster-calibrated judging. Detailed prototypes and trap-to-score mappings are deferred to the appendix.}
\label{fig:skill_prototype_compact}
\end{figure}

\subsection{Skill-Conditioned Reward and Optimization}
\label{sec:router_optimization}
\paragraph{Routing and Reward Evaluation.}To recover mid-level structure during training, we train a lightweight \emph{Router} (Qwen3-4B-Instruct-2507) to map raw trajectories into structured $\langle\text{subgoal, skill, step}\rangle$ triples. During RL, the Router assigns student-generated subgoals to their corresponding Skill Prototypes. A judge LLM then evaluates these subgoals with scores in $\{0, \dots, 3\}$ conditioned on the prototype's checklist. To ensure reward reliability, we adopt a calibration protocol that (i) verifies rewards across multiple prompt variants, (ii) leverages subgoal-level accuracy statistics, and (iii) monitors consistency via repeated evaluation on fixed anchor subgoals. The final process reward is a weighted average of these calibrated subgoal scores. Full details are provided in Appendix~\ref{appendix:reward_calibration}.
\paragraph{Policy Optimization with Adaptive Prototypes.}The student model is optimized using the GRPO objective based on these skill-conditioned rewards. To mitigate distributional shifts, we implement an \textbf{adaptive refresh} mechanism: every 1,000 training steps, we re-cluster accumulated trajectories to update the Skill Prototype library. This ensures that the supervision remains aligned with the model’s evolving reasoning patterns while maintaining stable mid-level abstractions for consistent credit assignment.

\section{Experimental Setup}

\paragraph{Experimental Settings.}
We conduct reinforcement learning experiments using Qwen3-4B-Instruct-2507%
\footnote{\url{https://huggingface.co/Qwen/Qwen3-4B-Instruct-2507}}
as the primary policy model, and additionally evaluate generality with LLaMA-3.2-3B-Instruct%
\footnote{\url{https://huggingface.co/meta-llama/Llama-3.2-3B-Instruct}}.
Process-level rewards are provided by an LLM-based judge (GPT-5-mini)\footnote{\url{https://platform.openai.com/docs/models/gpt-5-mini}}.
The Router is implemented as a lightweight model fine-tuned from the same base model as the policy.

Training is performed on approximately 10k problems sampled from the MATH dataset%
\footnote{\url{https://huggingface.co/datasets/open-r1/OpenR1-Math-220k/tree/main}}
and the ToolACE~\cite{liu2024toolace} dataset.
These data are used to construct skill annotations, train the Router, and perform GRPO-based policy optimization.
For mathematical reasoning tasks, we primarily rely on the Python interpreter, which is natively supported by Qwen3.

Applying the skill clustering protocol described in Sec.~\ref{prototype and cluster}, we obtain a compact mid-level skill space.
Initially, clustering yields 418 skill clusters for mathematical reasoning and 472 clusters for tool-use trajectories.
Clusters are periodically refreshed during training to incorporate newly observed trajectories.
After convergence, the skill space contains 424 clusters for mathematical reasoning and 503 clusters for BFCL-style tool-use tasks.

The Router is trained on judge-annotated trajectories with a 9:1 train--test split, achieving 98.6\% accuracy in skill prototype retrieval on held-out data.
Detailed subgoal-, skill-, and step-level evaluation metrics are reported in Appendix~\ref{app:router}.

During training, we combine process-level and final-answer rewards, with weights of 0.3 and 0.7 respectively.
Unless otherwise specified, this reward composition is used throughout.

The policy model is optimized using GRPO with skill-conditioned rewards.
We evaluate mathematical reasoning on MATH500 and AIME25, and assess tool-use generalization on BFCL v4~\cite{patilberkeley} (last updated: 2025-11-03) using the official evaluation scripts. Result are evaluated on latest BFCL V4. As a result, our reported numbers may differ from those reported by prior baselines evaluated on earlier versions of the benchmark.
We report pass@1, averaged over eight independent runs to reduce variance.
Additional training details and learning curves are provided in Appendix~\ref{app:training_details}.

\paragraph{Baselines.}
We compare against a step-level process reward (PRM) baseline that assigns rewards directly to individual reasoning steps using the same LLM-based judge, without mid-level skill abstraction.
We further include strong recent baselines for mathematical reasoning, including ReST~\cite{lin2025rest}, EGPO~\cite{hao2025reasoning}, and NPR~\cite{wu2025native}, following their official implementations.

\begin{table*}[htbp]
\centering
\small
\renewcommand{\arraystretch}{1.15}
\resizebox{1\textwidth}{!}{%
\begin{tabular}{c|cc|cccccc}
\toprule
\textbf{Method} 
& \multicolumn{2}{c|}{\textbf{MATH}} 
& \multicolumn{6}{c}{\textbf{BFCL}} \\

& MATH500 
& AIME25 
& Overall
& Web Search
& Memory
& Single (Live) 
& Single (Non-Live) 
& Multi \\
\midrule

\multicolumn{9}{l}{\textbf{Qwen3-4B-Instruct-2507}} \\
\midrule
Base     
& 89.1 & 43.3 & 33.0 & 5.0 & 12.7 & 86.7 & 74.5 & 19.4 \\
PRM      
& 92.3 & 51.7 & 44.6 & 10.0 & 15.2 & 88.5 & 77.6 & 28.6 \\
ReST     
& 88.5 & 43.3 & 38.9 & 5.0 & 13.1 & 87.2 & 76.5 & 25.3 \\
EGPO     
& 90.2 & 43.3 & 48.3 & 15.5 & 17.9 & 89.6 & 79.8 & 30.4 \\
NPR      
& 91.7 & 53.8 & 35.2 & 5.5 & 14.2 & 87.1 & 73.8 & 17.7 \\
\rowcolor{Highlight}
SCRIBE (Ours)     
& 95.8 & 63.3 & 51.3 & 12.5 & 15.8 & 90.0 & 80.5 & 33.3 \\
\midrule

\multicolumn{9}{l}{\textbf{LLaMA 3.2-3B-Instruct}} \\
\midrule
Base     
& 40.8 & 1.7 & 21.5 & 0.5 & 5.2 & 82.0 & 58.3 & 3.9 \\
PRM      
& 48.3 & 6.7 & 24.8 & 5.0 & 8.3 & 83.8 & 59.6 & 8.2 \\
ReST     
& 40.2 & 1.7 & 23.2 & 4.5 & 9.1 & 83.2 & 60.5 & 5.2 \\
EGPO     
& 55.7 & 6.7 & 28.2 & 6.5 & 9.9 & 84.2 & 61.1 & 9.1 \\
\rowcolor{Highlight}
SCRIBE (Ours)     
& 63.4 & 15.8 & 30.8 & 8.5 & 11.2 & 85.5 & 62.0 & 13.5 \\
\bottomrule
\end{tabular}%
}
\caption{Main results on mathematical reasoning and tool-use benchmarks. Result are evaluated on latest BFCL V4 (updated on Nov.03 2025, results might be various from origin reports.) }
\label{tab:main_results}
\end{table*}

\section{Main Result}
Table~\ref{tab:main_results} summarizes the main results on mathematical reasoning and general tool-use benchmarks.
Overall, our method consistently outperforms prior approaches across \emph{both} mathematical reasoning and tool-use settings, demonstrating its effectiveness beyond a single task domain.
On \textbf{Qwen3-4B-Instruct-2507}, our approach achieves a new best performance on MATH benchmarks, improving MATH500 accuracy from 92.3 (PRM) and 90.2 (EGPO) to 95.8, and AIME25 accuracy from 51.7 (PRM) to 63.3.
Notably, these gains in mathematical reasoning do not come at the cost of tool-use capability: on BFCL, our method attains the highest overall score of 51.3, outperforming strong baselines such as EGPO (48.3) and PRM (44.6), with consistent improvements across both single-step and multi-step tool-use scenarios.

In addition, we observe that even a \emph{simple step-level PRM baseline} remains competitive compared to more complex training schemes.
For example, PRM already yields substantial gains over the base model on both MATH500 (from 89.1 to 92.3) and BFCL Overall (from 33.0 to 44.6) for Qwen3.
However, our method further improves upon PRM by a large margin across all evaluated settings, including a +3.5 gain on MATH500 and a +6.7 gain on BFCL Overall.
A similar trend holds for \textbf{LLaMA~3.2-3B-Instruct}, where our approach improves MATH500 accuracy from 48.3 (PRM) to 63.4 and BFCL Overall from 24.8 to 30.8.
These results suggest that while PRM provides a strong and robust baseline, our prototype-conditioned reward formulation yields more consistent and transferable improvements across reasoning and tool-use tasks.

\section{Research Questions and  Ablation Studies}


\subsection{RQ1: Does Mid-Level Execution Improvement Lead to Emergent High-Level Planning Ability?}

\label{sec:rq1_dynamics}
A central question in this work is whether improving mid-level reasoning abilities can translate into gains in high-level planning.
While our training directly optimizes mid-level behavior via skill-conditioned rewards, it does not explicitly supervise high-level plan generation.

\paragraph{Structural training dynamics.}
To study whether mid-level improvements translate into high-level gains, we track \emph{structural training dynamics} at the subgoal and plan level throughout training, without relying on token-type annotation.

\emph{Mid-level execution} is evaluated by both subgoal success and reliability.
For each subgoal, we perform multiple independent rollouts and report \emph{Mid-level Success} ($\mathrm{MidSucc}$) as the macro-averaged completion rate over 64 trials.
To capture execution stability, we additionally report \emph{Mid-level Uncertainty} ($\mathrm{MidUnc}$), which measures the variability of outcomes across repeated rollouts, with lower values indicating more consistent execution.

\paragraph{High-level ability (HighLvl).}
We measure \emph{high-level planning ability} via \emph{execution-verified plan selection}.
For each task $x$, we collect a set of candidate subgoal sequences (plans) and determine their viability through empirical execution.
We define the model's preference score for a plan $\pi$ as its length-normalized log-probability:
\begin{equation}
r(\pi) = \frac{1}{|\pi|} \log p_\theta(\pi \mid x).
\end{equation}
The \emph{HighLvl} metric measures the fraction of viable--non-viable plan pairs for which the viable plan $\pi^+$ is ranked above the non-viable plan $\pi^-$ according to $r(\pi)$, which is equivalent to an AUC-style plan discrimination score.

To bridge mid-level execution and high-level plan selection, we additionally track \emph{plan separability} ($\mathrm{PlanSep}$), a structural signal that quantifies how distinctly viable and non-viable subgoal sequences can be distinguished based on their empirical execution outcomes.
When mid-level execution is unstable, different plans often fail for unrelated reasons, resulting in overlapping success distributions and low separability.
As execution becomes more reliable, successful and unsuccessful plans diverge more clearly, increasing $\mathrm{PlanSep}$ and enabling more effective plan-level discrimination.

Intuitively, as mid-level execution becomes more reliable and less variable, the empirical success distributions of viable and non-viable plans become increasingly separable.
This sharpened distinction facilitates more reliable plan selection, leading to improvements in $\mathrm{HighLvl}$ even without explicit supervision at the planning level.
Full metric definitions, labeling thresholds, execution protocols, and the formal definition of $\mathrm{PlanSep}$ are provided in Appendix~\ref{app:evps}. For tabular reporting, we aggregate math performance by macro-averaging results on MATH500 and AIME25,
as our goal here is to highlight cross-domain structural trends rather than dataset-specific effects.

\begin{figure}[t]
    \centering
    \includegraphics[width=\columnwidth]{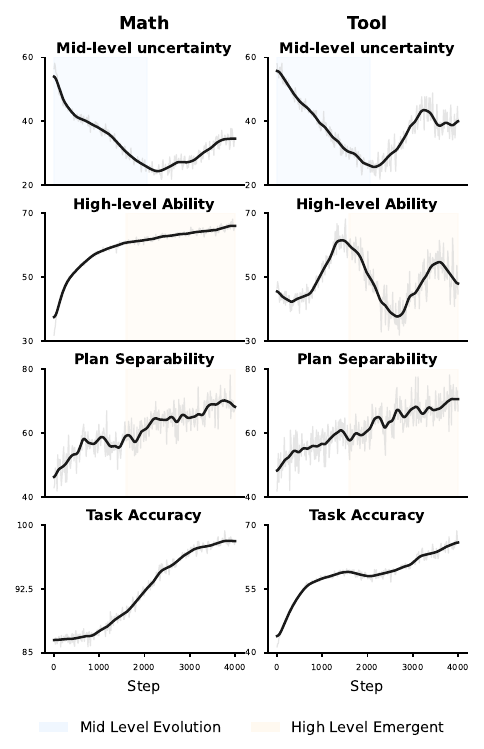}
    \caption{Structural training dynamics of our method.
    From top to bottom, we report mid-level uncertainty, high-level plan selection ability, plan separability, and final task accuracy.
    Results are shown separately for math (left column) and tool-use (right column) on Qwen3-4B-SCRIBE.
    }
    \label{fig:ab1_structural_dynamics}
\end{figure}

\begin{table}[t]
\centering
\small
\setlength{\tabcolsep}{3pt}
\renewcommand{\arraystretch}{1.15}
\begin{tabularx}{\columnwidth}{l|cc|cc}
\toprule
\textbf{Model} 
& \multicolumn{2}{c|}{\textbf{Mid-Level Success }} 
& \multicolumn{2}{c}{\textbf{High-Level Ability }} \\
& \textbf{Math} & \textbf{Tool} 
& \textbf{Math} & \textbf{Tool} \\
\midrule
\multicolumn{5}{l}{\textit{Qwen3-4B}} \\
\midrule
Base  & 78.2 & 46.0 & 85.1 & 68.7 \\
PRM   & 83.4 & 50.5 & 87.0 & 70.3 \\
\rowcolor{Highlight}
SCRIBE & 87.6 & 61.4 & 88.9 & 75.1 \\
\midrule
\multicolumn{5}{l}{\textit{LLaMA-3.2-3B}} \\
\midrule
Base  & 66.9 & 25.8 & 29.3 & 34.6 \\
PRM   & 72.3 & 30.4 & 34.7 & 36.0 \\
\rowcolor{Highlight}
SCRIBE & 75.6 & 33.2 & 41.0 & 42.2\\
\bottomrule
\end{tabularx}
\caption{Structural training dynamics across math and tool-use domains.}
\label{tab:ablation_mid_high_cross_domain}
\end{table}

\paragraph{Summary.}
Although our training explicitly targets mid-level execution via skill-conditioned rewards,
the resulting improvements are not confined to this level.
As shown in Fig.~\ref{fig:ab1_structural_dynamics}, stabilization of mid-level execution precedes a monotonic increase in plan separability
and a delayed but consistent rise in high-level plan selection ability, despite the absence of direct planning supervision.
Table~\ref{tab:ablation_mid_high_cross_domain} corroborates this trend across models and domains, where gains in mid-level success are consistently accompanied
by improvements in high-level ability.
Together, these results indicate that improving mid-level execution reliability can induce emergent gains in high-level planning
through enhanced structural separability rather than isolated local optimization.

\subsection{Can Mid-Level Skill Supervision Complement Low-Level Tool Optimization?}
\label{ab2}
Most existing approaches to tool-augmented reasoning focus on improving low-level execution,
such as increasing the reliability of tool invocation, argument formatting, or result parsing.
These methods primarily operate at the level of individual tool calls and aim to reduce execution failures.
In contrast, our approach targets mid-level skill execution by providing structured,
skill-conditioned supervision over reusable tool-using behaviors.
An important open question is whether improvements at these two levels are complementary,
or whether gains from mid-level supervision diminish once low-level execution is sufficiently optimized.

To study this interaction, we combine our method with a representative low-level tool optimization approach.
Specifically, we adopt FunRL~\cite{hao2025reasoning}, a reinforcement learning method designed to improve
function-calling reliability and argument correctness at the execution level.
We train FunRL directly on the \textsc{LLaMA-3.2-3B} backbone to obtain a backbone-aligned low-level baseline,
ensuring a controlled comparison without confounding architectural differences.

We compare four settings: the base model, a low-level optimized model trained with FunRL alone,
a model trained with mid-level skill supervision alone, and a combined setting that integrates both.
This design allows us to isolate the individual contributions of low- and mid-level optimization,
and to assess whether mid-level supervision provides additive benefits beyond improved low-level execution.
\begin{table}[t]
\centering
\small
\renewcommand{\arraystretch}{1.15}
\setlength{\tabcolsep}{6pt}
\begin{tabular}{l|c}
\toprule
\textbf{Setting} & \textbf{Tool-Use (\%)} \\
\midrule
Base Model                          & 21.5 \\
Low-Level Only (FunRL)              & 25.2 \\
Mid-Level Only (SCRIBE)               & 30.8 \\
\rowcolor{Highlight}
Low+ Mid-Level (FunRL + SCRIBE)& \textbf{33.4} \\
\bottomrule
\end{tabular}
\caption{
Effect of combining low-level tool optimization with mid-level skill-based supervision
on \textsc{LLaMA-3.2-3B}.
While FunRL improves execution-level reliability, integrating mid-level supervision
yields additional gains, indicating complementary benefits across optimization levels.
}
\label{tab:low_mid_ablation}
\end{table}

\paragraph{Conclusion.}
Low-level tool optimization and mid-level skill supervision act on distinct and non-conflicting aspects of tool-augmented reasoning.
While FunRL improves execution-level reliability (21.5$\rightarrow$25.2) and mid-level supervision independently yields larger gains (21.5$\rightarrow$30.8),
their combination leads to further improvements (25.2$\rightarrow$33.4).
These results indicate an additive and complementary relationship, where mid-level supervision augments rather than interferes with low-level optimization.

\subsection{Ablation on Reward Weighting.}
We study the effect of balancing process-level and outcome-level rewards.
Specifically, we vary the weight of the process reward $w_p \in \{0.1, 0.3, 0.5, 0.7\}$, with the outcome reward weight set to $1 - w_p$.
All other training settings are kept fixed.
Performance is evaluated on MATH500, AIME25, and BFCL v4 (Overall). Table~\ref{tab:reward_weight_ablation} shows that the choice of reward weighting substantially affects both mathematical reasoning and tool-use performance.
In particular, assigning a moderate weight to the process-level reward ($w_p=0.3$) achieves the best overall trade-off across MATH500, AIME25, and BFCL v4, whereas excessive emphasis on either intermediate or final rewards leads to performance degradation.

\begin{table}[t]
\centering
\small
\renewcommand{\arraystretch}{1.15}
\begin{tabular}{c|ccc}
\toprule
\textbf{$w_p$} 
& \textbf{MATH500} 
& \textbf{AIME25} 
& \textbf{BFCL (Overall)} \\
\midrule
0.1 & 92.1 & 51.7 & 33.3 \\
\rowcolor{Highlight}
0.3 & 95.8 & 63.3 & 51.3 \\
0.5 & 91.0 & 51.7 & 36.1 \\
0.7 & 82.7 & 33.4 & 29.9 \\
\bottomrule
\end{tabular}
\caption{Ablation study on the weighting between process-level and outcome-level rewards, on Qwen3-4B-Instruct-2507.
The outcome reward weight is set to $1 - w_p$.}
\label{tab:reward_weight_ablation}
\end{table}

\section{Conclusion}

We introduced \textsc{SCRIBE}, a framework that enhances tool-augmented reasoning by intervening at a novel \textbf{mid-level abstraction}. By anchoring process rewards in a library of \textit{Skill Prototypes}, \textsc{SCRIBE} transforms subjective LLM judging into grounded, diagnostic verification. Our results demonstrate state-of-the-art performance, notably improving the \textbf{AIME25} score of a 4B model from \textbf{43.3\% to 63.3\%} and doubling success rates in complex multi-turn tool use. 

Beyond performance, our analysis of training dynamics characterizes a \textbf{co-evolutionary principle}: mid-level skill mastery acts as a necessary precursor to the emergence of strategic high-level planning. We further show that \textsc{SCRIBE} is highly \textbf{additive}, complementing low-level tool optimizations to yield superior synergistic effects. Future work will explore more autonomous ways to evolve skill libraries in real-time to handle open-domain tasks. Additionally, we aim to investigate how multi-level hierarchical supervision can be unified into a single objective to further enhance the strategic depth of autonomous agents.

\section{Limitations}
Although our approach yields consistent improvements on mathematical reasoning and tool-use benchmarks, several limitations remain.
First, our experiments focus on small- to mid-sized instruction-tuned models; it is unclear how the proposed skill-conditioned supervision scales to substantially larger architectures or to models trained with different post-training pipelines.
Second, the construction of skill prototypes and routing relies on clustering heuristics and judge-annotated data, which may introduce biases or limit adaptability in domains with highly diverse or ill-defined skill boundaries.
Finally, our evaluation primarily targets structured reasoning and tool-use tasks; the effectiveness of the approach for open-ended generation or non-instrumental reasoning remains an open question.

\section{Ethics}
This work relies on publicly available datasets and uses an LLM-based judge accessed via the OpenAI API for reward evaluation~\cite{openai_api}.
We do not access, attempt to access, or infer any proprietary training data or internal components of the underlying models.
All experiments are conducted using standard model inference and optimization procedures, without collecting or processing personal or sensitive user data.

\paragraph{Risks}
The datasets used in our experiments are sourced from publicly available benchmarks and may contain unintended biases, errors, or harmful language.
While our method aims to improve the consistency of reward signals during training, it could potentially amplify biases present in the judge or training data if deployed without additional safeguards.
We also use GPT5 to assist with minor grammatical corrections in this paper.

\bibliography{custom}

\appendix

\section{Reward Calibration and Consistency Analysis}
\label{appendix:reward_calibration}

\subsection{Prompt Variants}
\paragraph{Prompt Variant M1 (Narrative Decomposition).}
You are given a complete solution to a math problem.
Analyze the solution and identify the key intermediate subgoals that are required to reach the final answer.
For each subgoal, briefly describe the mathematical objective being achieved and the primary reasoning skill involved.
Focus on meaningful reasoning stages rather than low-level algebraic steps.
The subgoals should be ordered and together sufficient to solve the problem.\\

(Example output omitted.)

\paragraph{Prompt Variant M2 (Coverage-Constrained Segmentation).}
You will receive a solution trajectory for a math problem.
Segment the entire solution into a sequence of subgoals such that every part of the trajectory belongs to exactly one subgoal.
Each subgoal should correspond to a coherent reasoning objective.
For each subgoal, identify the dominant mathematical skill applied.
Ensure no overlap or omission across subgoals.\\

(Example output omitted.)\\
\paragraph{Results.}
Across math problems, both prompt variants recover highly consistent subgoal structures.
The ordering and skill attribution of subgoals largely agree with those produced by the main prompt, with high rank-level agreement in reward scores.
This indicates that subgoal extraction and subsequent reward evaluation are not sensitive to prompt phrasing.
\paragraph{Prompt Variant T1 (Intent-Oriented Decomposition).}
You are given a multi-turn, tool-using interaction.
Identify the sequence of subgoals that the system completes in order to satisfy the user's request.
Each subgoal should describe a concrete intent or intermediate objective and the primary capability required to achieve it.
Subgoals may span multiple turns and should reflect mid-level planning rather than surface actions.

(Example output omitted.)
\paragraph{Prompt Variant T2 (Failure-Aware Segmentation).}
Given a tool-using conversation trajectory, segment it into subgoals such that the entire interaction is covered.
If the trajectory includes partial failures, corrections, or fallback behaviors, include them as explicit subgoals.
For each subgoal, identify the dominant tool-use or reasoning skill involved.

(Example output omitted.)
\paragraph{Results.}
For tool-using tasks, both prompt variants produce subgoal partitions that closely match the main prompt.
Reward scores assigned based on these subgoals remain stable across prompts, with only minor variations in boundary placement for a small number of tool-related steps.
Overall, the judge exhibits robust agreement across prompt formulations.
\subsection{Subgoal-Level Reward Calibration}

The LLM-based judge assigns discrete reward scores in $\{0,1,2,3\}$ to each subgoal.
To improve alignment between reward magnitude and actual correctness, we calibrate rewards using empirical subgoal-level outcome statistics.

For each subgoal type, we estimate the empirical success rate of final task completion conditioned on the assigned reward score.
If a lower raw reward (e.g., 2) corresponds to a success rate comparable to a higher reward (e.g., 3), the reward is calibrated upward.
Conversely, rewards that systematically overestimate correctness are adjusted downward.
Calibration is implemented via a subgoal-specific lookup table, preserving relative ordering while improving reward validity.
\paragraph{Human Agreement.}
We uniformly sample 100 subgoal instances across all reward levels and manually verify whether the assigned rewards reasonably reflect subgoal quality.
No systematic disagreement is observed, providing a sanity check that the reward scores align with human judgment.
\subsection{Reward Consistency via Anchor Subgoals}

To assess reward consistency over time, we select 20 fixed anchor subgoals for each task type.
Each anchor subgoal is re-evaluated five times at different checkpoints during training.

The assigned rewards are highly stable.
Only two anchor instances in tool-use tasks receive slightly different scores across evaluations, while all remaining anchor subgoals retain identical rewards.
This suggests that the judge’s evaluation criteria remain consistent over time without noticeable drift.

\section{Router Evaluation}
\label{app:router}

We evaluate the Router on held-out judge-annotated trajectories to assess its ability to recover mid-level reasoning structure.
The annotated data are split into training and test sets with a 9:1 ratio.
Evaluation is performed on the test split and focuses on three aspects: step segmentation, skill prediction, and Skill Prototype retrieval.

The Router takes as input a problem and its corresponding raw reasoning trajectory, and outputs an ordered sequence of $\langle$subgoal, skill, step$\rangle$ tuples together with the associated Skill Prototype.
All reported metrics are computed by comparing the Router outputs against judge-provided annotations.
\paragraph{Step Segmentation Accuracy.}
We evaluate step prediction using span-level exact match (EM).
A predicted step is considered correct if its start and end positions exactly match the annotated step span in the original trajectory.
This metric directly measures whether the Router correctly partitions the trajectory into non-overlapping reasoning segments.
\paragraph{Skill Prediction Accuracy.}
For each predicted subgoal, we evaluate whether the associated skill label matches the annotated skill.
We report classification accuracy over all subgoals in the test set.
\paragraph{Skill Prototype Retrieval Accuracy.}
We evaluate whether each subgoal is routed to the correct Skill Prototype.
A prediction is considered correct if the retrieved prototype matches the annotated prototype.
This metric reflects the Router’s effectiveness in providing correct skill-level context for reward evaluation.
\begin{table}[htbp]
\centering
\small
\begin{tabular}{l|c}
\toprule
\textbf{Metric} & \textbf{Accuracy (\%)} \\
\midrule
Step Segmentation (EM)        & 94.6 \\
Skill Prediction              & 95.3 \\
Skill Prototype Retrieval     & 98.6 \\
\bottomrule
\end{tabular}
\caption{Router evaluation results on held-out annotated trajectories.
Step segmentation is measured by span-level exact match, while skill prediction and prototype retrieval are measured by classification accuracy.}
\label{tab:router_eval}
\end{table}

\section{Additional Training Details}
\label{app:training_details}

We provide additional details on GRPO training and stability diagnostics to facilitate reproducibility.

\paragraph{GRPO Configuration.}
The student model is optimized using GRPO with skill-conditioned process-level rewards.
For mathematical reasoning, we sample 8 rollouts per problem over 10k training problems.
We include a KL regularization term to constrain policy updates and an entropy bonus to encourage exploration.
Reward shaping is applied to normalize process-level scores.
Unless otherwise specified, the same optimization settings are used across all methods.

\paragraph{Optimization Settings.}
We use a fixed learning rate and batch size throughout training, with a local batch size of 128.
All hyperparameters are selected from standard ranges used in prior GRPO and RLHF-style training and are kept constant across baselines.

\paragraph{Training Stability.}
To monitor training stability, we track reward statistics over time, including mean reward and reward variance.
We observe no evidence of reward collapse during training.

\onecolumn

\section{Additional Related Works on LLMS Reasoning Capacity}
Among various LLM abilities, foundational research in semantic matching~\cite{xue2023dual, xue2025structcoh} and temporal QA~\cite{xue2024qa} has paved the way for complex understanding. While domain adaptation~\cite{huang2023dtbs} and test-time integration~\cite{huang2025cosmic} enhance robustness, reasoning ability achieved by post-training~\cite{huang2026rlvr, lan2025mappo} stands as a key frontier. Given that traditional SFT may suffer from incomplete learning~\cite{xue2026sft}, recent works introduce uncertainty-aware reward modeling~\cite{xue2026reward}, dense feedback from LM critics~\cite{cao2024enhancing}, and preference-driven code generation~\cite{li2025preference} to refine alignment. Moreover, recent studies have advanced role-playing reasoning through human-actor emulation~\cite{chen2026actormindemulatinghumanactor} and improved continual learning via prototype-conditioned generative replay~\cite{chen-zeng-2025-prototype}.

The scope of reasoning has expanded into multimodal domains, where models now "think with 3D"~\cite{chen2025think} and utilize reinforced visual perception~\cite{chen2025visrl, chen2025sifthinker}. Advanced frameworks like OmniVideo-R1~\cite{chen2026omnivideo} and DV-Matcher~\cite{chen2025dv} further integrate audio-visual and geometric cues. Such multimodal intelligence significantly boosts retrieval performance, including chat-driven search~\cite{xie2025chat, xie2026conquer} and vision-driven video representation~\cite{xie2026hvd, xie2026delving}. Furthermore, safeguarding techniques~\cite{yu2024robust, lan2025contextual} and hallucination mitigation~\cite{yu2024mechanistic, ji2025calibrating} ensure the reliability of these complex systems.

To support deployment, system efficiency and continual learning have become indispensable. Inference is optimized via multi-tier KV cache management~\cite{chu2025mcam}, selective sharing~\cite{chu2025selective}, and adaptive MoE scheduling~\cite{chu2025dynamic, shen2025expertflow}. Moreover, lightweight techniques like parameter-efficient sampling~\cite{yao2024swift}, structured pruning~\cite{li2026sepprune}, and distillation~\cite{li2025frequency} enable efficient task execution, from medical imaging~\cite{Li2025Efficient} to web agents~\cite{ding2026dynaweb}. Finally, reversible alignment~\cite{xiao2026reversible} and unified LoRA generation~\cite{xiaometa, chen2024three} allow models to learn continually, while comprehensive interaction surveys~\cite{li2026comprehensive} and robust federated learning~\cite{chen2024confusion} continue to push the boundaries of scalable AI.

Beyond core architecture, recent research emphasizes a data-centric perspective on the entire LLM lifecycle~\cite{rao2025data}, where mathematical reasoning is further enhanced by adversarial data synthesis~\cite{yu2026mathagent} and iterative DPO with dynamic sampling~\cite{rao2025dynamic}. To improve model robustness and learning efficiency, advanced techniques such as difficulty-based sample weighting~\cite{zhou2022understanding}, implicit data augmentation~\cite{zhou2024boosting}, and adversarial training with anti-adversaries~\cite{zhou2024adversarial} have been introduced.

In the context of embodied AI and spatio-temporal reasoning, unified token spaces for human-object interaction~\cite{yang2026unihoi} and object-centric decoupling for robotic manipulation~\cite{yang2026instrucrobo} have improved explainable decision-making. These abilities are further integrated into 3D scene understanding~\cite{yang2026unibvr} and autonomous driving through spatio-temporal Chain-of-Thought (CoT)~\cite{zeng2025fsdrive} and dual-memory vision-language navigation~\cite{zeng2025janusvln}. Furthermore, the application of multi-scale graph learning and Transformers has shown significant potential in predicting complex real-world dynamics, including financial fraud detection~\cite{lei2026multiscale}, traffic flow prediction~\cite{shen2026mftformer}, disaster risk assessment~\cite{shen2025aienhanced}, mental health~\cite{yang2026exploring}, recommendations~\cite{wang2025improving} and socio-economic metrics such as labor market gradients~\cite{liu2026health} and business resilience indices~\cite{Shen2026bri}.
\section{Prompts for Subgoal and step decomposition}
\label{full prompt}
We use structured prompts to decompose each problem or trajectory into mid-level subgoals and associated skills, which serve as the basis for our skill-aware supervision and analysis.

\begin{tcolorbox}[promptbox, title=Subgoal \& Skill Extraction Prompt (Math)]
{\slshape
I will give you a math problem.Break the solution into several subgoals, where completing these subgoals in order is sufficient to solve the whole problem. For each subgoal, explicitly name the main skill required, and give one short step showing how it is applied.

Use this format:
(1) subgoal: <what this subgoal accomplishes>
    skill: <main skill needed>
    step: <one short sentence>

Example:
\noindent
\textbf{subgoal:} maximize $u = (x+4)y$ subject to $x^2 + y^2 = 1$ \\
\textbf{skill:} optimization with constraint (Lagrange multipliers or trigonometric substitution) \\
\textbf{step:} set $x = \cos \theta$, $y = \sin \theta$, so
\[
u(\theta) = (\cos \theta + 4)\sin \theta
           = \tfrac{1}{2}\sin 2\theta + 4\sin \theta,
\]
and maximize over $\theta$.

Now apply this style to the following problem: 
}
\end{tcolorbox}

\begin{tcolorbox}[promptbox, title=Example Output (Tool-Using Trajectory), listing only]
{\slshape
You will receive a multi-turn, tool-using conversation trajectory. The user's final goal is achieved step by step through the completion of multiple subgoals. Subgoals may span multiple turns and should reflect MID-LEVEL reasoning or planning.

Your task is to identify all subgoals that are carried out in order to accomplish the user's final goal.

For each subgoal:
Specify the primary skill required to complete this subgoal.
Identify the corresponding step(s) in the original trajectory.Each subgoal must correspond to a single contiguous, unmodified subpart of the original trajectory.The entire trajectory must be partitioned into such subparts with no overlap and no omission:every part of the trajectory should belong to exactly one subgoal.

If the task is partially completed or fails, include subgoals that capture request decomposition,
 recognition of capability limitations, or fallback handling.

Example output:
(1) subgoal: get the top market trend in us
    skill: situational assessment from structured signals
    step: Here are the top Market Trends in the US right now:1. **S\&P 500**: The Standard \& Poor's 500 Index is a market-capitalization-weighted index of the 500 largest U.S. publicly traded companies. Its current value is 4172.80 with a percentage change of +0.68\%.\\2. **DOW J**: The Dow Jones Industrial Average is a price-weighted average of 30 blue-chip stocks that are generally the leaders in their industry. Its current value is 34479.60 with a percentage change of +0.47\%.\\3. **NASDAQ**: The NASDAQ Composite is a broad-based capitalization-weighted index of stocks in all three NASDAQ tiers: Global Select, Global Market and Capital Market. Its current value is 13691.30 with a percentage change of +0.90\%.This information can help you make informed decisions about your investment plans.
}
\end{tcolorbox}

\section{Execution-Verified Plan Selection and Structural Metrics}
\label{app:evps}

\paragraph{Mid-level success and uncertainty.}
For each subgoal $u$, we sample $R$ independent rollouts and define the subgoal success rate as
\begin{equation}
\mathrm{MidSucc}(u)
= \frac{1}{R}\sum_{r=1}^{R} \mathbb{I}\!\left[\text{rollout } r \text{ successfully completes } u\right],
\end{equation}
where $R=64$ in all experiments.
We report the macro-average
$\mathrm{MidSucc}=\frac{1}{|U|}\sum_{u\in U}\mathrm{MidSucc}(u)$.

To quantify execution reliability, we define the mid-level uncertainty for subgoal $u$ as
\begin{equation}
\mathrm{MidUnc}(u)
= \mathrm{MidSucc}(u)\bigl(1-\mathrm{MidSucc}(u)\bigr),
\end{equation}
and report its macro-average
$\mathrm{MidUnc}=\frac{1}{|U|}\sum_{u\in U}\mathrm{MidUnc}(u)$.
This measure captures the variability of execution outcomes across repeated rollouts under a fixed subgoal specification.

\paragraph{Candidate plan construction and execution.}
For each task instance $x$, we collect a candidate plan set $\mathcal{P}(x)$ consisting of subgoal sequences extracted from prior rollouts.
These candidate plans are held fixed across training checkpoints.
Each plan $\pi\in\mathcal{P}(x)$ is executed for $T$ independent trials, yielding an empirical success rate
\begin{equation}
\hat{s}(\pi)
= \frac{1}{T}\sum_{t=1}^{T} \mathbb{I}\!\left[\text{execution of } \pi \text{ solves } x\right],
\end{equation}
with $T=5$ in all experiments.

\paragraph{Plan viability labeling.}
Based on empirical success rates, plans are partitioned into \emph{viable} and \emph{non-viable} sets.
A plan $\pi$ is labeled viable if $\hat{s}(\pi)\ge\tau_{\mathrm{hi}}$ and non-viable if $\hat{s}(\pi)\le\tau_{\mathrm{lo}}$; intermediate cases are discarded.
Unless otherwise specified, we use $\tau_{\mathrm{hi}}=0.5$ and $\tau_{\mathrm{lo}}=0.0$.
We verify that alternative rank-based labeling strategies (e.g., top-$m$ vs.\ bottom-$m$ plans) yield consistent qualitative trends.

\paragraph{Execution-verified plan selection}
Let $\mathcal{V}(x)$ and $\mathcal{N}(x)$ denote the sets of empirically viable and non-viable plans for task $x$, respectively.
Given a preference score $r(\pi)$, we define
\begin{equation}
\mathrm{HighLvl}
= \mathbb{E}_{x \sim \mathcal{D}}
\left[
\frac{1}{|\mathcal{V}(x)||\mathcal{N}(x)|}
\sum_{\pi^+ \in \mathcal{V}(x)} \sum_{\pi^- \in \mathcal{N}(x)}
\mathbb{I}\!\left[r(\pi^+) > r(\pi^-)\right]
\right].
\end{equation}
This metric is equivalent to the AUC of a binary classifier that ranks viable versus non-viable plans using $r(\pi)$.

\paragraph{Plan separability.}
To analyze the structural conditions under which high-level plan selection improves, we additionally compute \emph{plan separability} ($\mathrm{PlanSep}$).
Unlike $\mathrm{HighLvl}$, which depends on model preferences, $\mathrm{PlanSep}$ is defined purely in terms of empirical execution outcomes.
Specifically, for each task $x$, we compute the gap between the mean execution success rates of viable and non-viable plans:
\begin{equation}
\mathrm{PlanSep}
=
\mathbb{E}_{x}\!\left[
\mathbb{E}_{\pi\in\mathcal{V}(x)} \hat{s}(\pi)
-
\mathbb{E}_{\pi\in\mathcal{N}(x)} \hat{s}(\pi)
\right].
\end{equation}
A larger value indicates that successful and unsuccessful plans are more clearly separated in execution outcome space, reflecting a structural signal induced by improved execution reliability.

\paragraph{Evaluation scale.}
For mid-level evaluation, we sample subgoals extracted from a fixed subset of problems in each dataset.
Specifically, we evaluate on $N_{\text{math}}$ problems from MATH500 and $N_{\text{aime}}$ problems from AIME25, yielding an average of approximately $S$ subgoals per problem.
Each subgoal is evaluated with $R{=}64$ independent rollouts.

For high-level evaluation, candidate plan sets are constructed from the same problem subset.
For each problem, we collect $K$ candidate subgoal sequences from prior rollouts (typically $K{=}5$--$7$).
Each candidate plan is executed for $T{=}5$ trials to estimate empirical viability.
High-level ability and plan separability are then computed by aggregating execution-verified statistics across all evaluated problems.

\section{Cluster and Prototype examples}

\label{Prototype}

\paragraph{Math Prototype}

\begin{figure*}[t]
\begin{tcolorbox}[
    enhanced,
    colback=gray!5,
    colframe=gray!70!black,
    fonttitle=\bfseries,
    title={(\uppercase\expandafter{\romannumeral 3}) Skill Prototype (Mathematical Reasoning)},
    width=\linewidth,
    breakable,
    sharp corners
]
\textbf{Skill Name:} Bound-Based Conclusion and Synthesis

\medskip
\textbf{Knowledge Scope:}
Upper/lower bounding; extremal principles; attainability (tightness) arguments; case coverage; domain constraints; logical implication.

\medskip
\textbf{Applicable Scenario:}
Subgoals that \emph{terminate} a reasoning chain by drawing a final conclusion from previously derived constraints (e.g., impossibility via bounds, identifying a maximal value, selecting a correct option after logical elimination), including cases where equality/tightness or feasibility must be justified.

\medskip
\textbf{Canonical Reasoning Pattern:}
\begin{enumerate}
    \item Identify relevant intermediate results (bounds, constraints) that the conclusion must follow from.
    \item Determine if the subgoal requires an implication or a tightness claim (e.g., "max equals bound").
    \item If concluding an extremum, check \textbf{tightness}: reference a construction or equality condition.
    \item Verify domain restrictions and case completeness (e.g., integer ranges, geometric feasibility).
    \item State the final conclusion succinctly (value, choice, or maximal attainable level).
\end{enumerate}

\medskip
\textbf{Judging Rubric (0--3 Step Score):}
\begin{itemize}
    \item \textbf{3 (Correct \& Complete):} Conclusion follows rigorously from prior results; addresses tightness/feasibility.
    \item \textbf{2 (Minor Flaw):} Correct strategy and conclusion, but with minor slips (arithmetic/wording) or obvious-but-unstated justifications.
    \item \textbf{1 (Major Logical Gap):} Correct skill type, but substantial omissions (e.g., ignores a case) making the conclusion unsupported.
    \item \textbf{0 (Wrong Skill):} Incompatible/premature conclusion or complete misapplication of bounding logic.
\end{itemize}

\medskip
\textbf{Common Traps \& Scoring Mapping :}
\begin{center}
\small
\begin{tabular}{|p{0.32\linewidth}|c|p{0.53\linewidth}|}
\hline
\textbf{Common Trap} & \textbf{Score} & \textbf{Reasoning for Penalty} \\ \hline
\textbf{Boundary Leak} (e.g., leaving $c < 44.75$ without floor) & \textbf{2} & Method is correct, but the final conclusion lacks the discrete precision required by the domain. \\ \hline
\textbf{Implicit Tightness} (Assuming max is the bound without proof) & \textbf{1} & A major logical gap; in competition math, a bound is not a maximum until attainability is shown. \\ \hline

 \hline
\end{tabular}
\end{center}

\medskip
\textbf{Representative Reference Step:} 
From "for $n \ge 5$, $6n \le 2^n$ (impossible)" and "levels $n \le 4$ are feasible", conclude the max level is 4.(more ommitted)
\end{tcolorbox}
\end{figure*}

\paragraph{Tool use Prototype}
\begin{figure*}[t]
\begin{tcolorbox}[
    enhanced,
    colback=gray!5,
    colframe=gray!70!black,
    fonttitle=\bfseries,
    title={(\uppercase\expandafter{\romannumeral 4}) Skill Prototype (Tool-Using Reasoning)},
    width=\linewidth,
    breakable,
    sharp corners
]
\setlength{\baselineskip}{0.95\baselineskip}
\setlength{\parskip}{0pt}

\textbf{Skill Name:} Capability Limitation Handling and Tool-Mediated Fallback Guidance

\vspace{4pt}
\textbf{Knowledge Scope:}
Tool selection and parameterization; structured API invocation; tool output parsing; user-facing summarization; uncertainty calibration; data access limitations.

\vspace{4pt}
\textbf{Applicable Scenario:}
Tool-using tasks where the agent must (i) invoke a tool for analysis or retrieval, (ii) interpret and present tool outputs, and/or (iii) recognize missing access, unavailable data, or capability limits and provide actionable fallback guidance (e.g., how to obtain required inputs, what can be done instead, or how the agent can proceed once inputs are provided).

\vspace{4pt}
\textbf{Canonical Reasoning Pattern:}
\begin{enumerate}
    \setlength{\itemsep}{2pt}
    \setlength{\parskip}{0pt}
    \setlength{\parsep}{0pt}
    \item \textbf{Goal \& Constraints Identification:} Restate the user intent and identify required inputs (entity IDs, time range, text to analyze) and constraints (access, tool availability, policy limits).
    \item \textbf{Tool Selection \& Invocation:} Choose the correct tool and issue a well-formed call with task-relevant parameters (e.g., \texttt{shareuid}, \texttt{from/to}; \texttt{text=}).
    \item \textbf{Structured Output Interpretation:} Parse returned fields and map them to the user request (e.g., \texttt{roa\_ratio} $\rightarrow$ ROA for FY2025; sentiment label + keywords).
    \item \textbf{Limitation Detection \& Fallback:} If the request cannot be completed (missing access/data/tool), explicitly state the limitation, offer concrete next steps (how to obtain the data / what to provide), and propose safe alternatives.
\end{enumerate}

\vspace{4pt}
\textbf{Judging Rubric (0--3 Step Score):}
\begin{itemize}
    \setlength{\itemsep}{2pt}
    \setlength{\parskip}{0pt}
    \setlength{\parsep}{0pt}
    \item \textbf{3 (Correct \& Complete):} Correctly identifies constraints...(ommitted)
    \item \textbf{2 (Minor Flaw):} Overall correct skill and intent, but with minor issues (slightly suboptimal parameters, small formatting problems, or partially specified fallback that is still usable).
    \item \textbf{1 (Major Logical Gap):} Correct general direction (tool use or limitation handling) but substantial problems: wrong/missing critical parameters, misreading key fields, vague or non-actionable fallback, or mixing speculation with tool outputs.
    \item \textbf{0 (Wrong Skill / Unsafe / Hallucination):} Uses the wrong tool or fabricates tool outputs/data, claims completion despite missing access, or provides unsafe/irrelevant guidance.
\end{itemize}

\vspace{4pt}
\textbf{Common Traps \& Scoring Mapping:}
\begin{center}
\small
\setlength{\tabcolsep}{4pt}
\renewcommand{\arraystretch}{1.05}
\begin{tabular}{|p{0.32\linewidth}|c|p{0.53\linewidth}|}
\hline
\textbf{Common Trap} & \textbf{Score} & \textbf{Reasoning for Penalty} \\ \hline
\textbf{Hallucinated Tool Output} (inventing data without a tool result) & \textbf{0} & Breaks tool grounding: the output is unverifiable and violates the tool-mediated protocol. \\ \hline
\textbf{Non-Actionable Limitation} (only says ``I can't'' without next steps) & \textbf{1} & Recognizes a limitation but does not complete the subgoal of providing feasible fallback guidance. \\ \hline
\textbf{Over-Refusal Despite Feasible Alternative} (refuses when public steps or user-provided inputs could proceed) & \textbf{2} & Limitation acknowledged, but the response underutilizes available actions; still partially helpful. \\ \hline
\end{tabular}
\end{center}

\vspace{4pt}
\textbf{Representative Reference Step:}
If a user requests ROA for \texttt{shareuid=6789} for FY2025, invoke \texttt{Financial Fundamentals API(shareuid=6789, from="2025-01-01", to="2025-12-31")}, then report ...(ommitted)
(more ommitted)
\end{tcolorbox}
\end{figure*}

\end{document}